\pdfoutput=1

\documentclass[11pt]{article}

\usepackage[final]{acl}

\usepackage{times}
\usepackage{latexsym}
\usepackage{url}
\usepackage[T1]{fontenc}
\usepackage{tcolorbox}

\usepackage[utf8]{inputenc}

\usepackage{microtype}
\usepackage{booktabs}
\usepackage{amssymb}
\usepackage{amsmath}
\usepackage{tabularx}
\usepackage{hyperref}
\usepackage{amsmath}
\usepackage{subfig}
\usepackage{pifont}
\usepackage{graphicx}
\usepackage{multirow}

\usepackage{inconsolata}
\definecolor{darkgreen}{rgb}{0.0, 0.5, 0.0} 
\definecolor{darkred}{rgb}{0.5, 0.0, 0.0}   

\newcommand{\cmark}{\textcolor{darkgreen}{\ding{51}}}%
\newcommand{\xmark}{\textcolor{red}{\ding{55}}}%
\title{Beyond the Turn-Based Game: Enabling Real-Time \\ Conversations with Duplex Models}

\author{Xinrong Zhang$^1$, Yingfa Chen$^1$, Shengding Hu$^1$, Xu Han$^1$$^{*}$, Zihang Xu$^1$, Yuanwei Xu$^3$\\
\textbf{Weilin Zhao$^1$, Zhiyuan Liu$^2$$^,$$^1$\thanks{Corresponding author: Xu Han and Zhiyuan Liu}, Maosong Sun$^1$}\\
{$^1$NLP Group, DCST, IAI, BNRIST, Tsinghua University, Beijing, China.}\\
{$^2$Quan Cheng Laboratory, Jinan, China.}\\
{$^3$Modelbest Inc.}\\
\texttt{zxr19@tsinghua.org.cn},
\texttt{\{hanxu2022,liuzy\}@tsinghua.edu.cn}
}

\begin{document}
\maketitle
\begin{abstract}
As large language models (LLMs) increasingly permeate daily lives, there is a growing demand for real-time interactions that mirror human conversations. 
Traditional turn-based chat systems driven by LLMs prevent users from verbally interacting with the system while generating responses.
To overcome these limitations, we adapt existing LLMs to \textit{duplex models} so that they can listen to users while generating output and dynamically adjust themselves to provide instant feedback. Specifically, we divide the queries and responses of conversations into several time slices and then adopt a time-division-multiplexing (TDM) encoding-decoding strategy to process these slices pseudo-simultaneously.
Furthermore, to make LLMs proficient enough to handle real-time conversations, we build a fine-tuning dataset consisting of alternating time slices of queries and responses and covering typical feedback types in instantaneous interactions.
Our experiments show that although the queries and responses of conversations are segmented into incomplete slices for processing, LLMs can preserve their original performance on standard benchmarks with a few fine-tuning steps on our dataset. 
Automatic and human evaluation indicate that duplex models make user-AI interactions more natural and human-like, and greatly improve user satisfaction compared to vanilla LLMs. 
Our duplex model and dataset are released~\footnote{code: \url{https://github.com/thunlp/duplex-model}; dataset: \url{https://huggingface.co/datasets/xinrongzhang2022/Duplex-UltraChat}}.

\end{abstract}

\section{Introduction}

Large language models (LLMs) have demonstrated impressive capabilities in various scenarios~\citep{chatgpt,gpt4,touvron2023llama,team2023gemini}. 
These large models are deeply integrated into our daily lives. Their extraordinary capabilities can satisfy users in many applications, such as coding assistants~\citep{chen2021evaluating,copilot,copilot-chat,copilot-voice,Rozire2023CodeLO,Li2023StarCoderMT}, task assistants~\citep{wang2023voyager,qian2023communicative,gpt-4o}, virtual role play~\citep{shao2023character,shanahan2023role}, and even emotional companions~\citep{CHATURVEDI2023122634,guingrich2023chatbots,pentina2023exploring}. 
 
Despite ongoing advancements, interactions with LLMs often fail to provide users with human-like interaction experience~\cite{hill2015real,mou2017media,zhou2023talking}. One reason is the turn-based nature of current chatbot implementations~\citep{skantze2021turn}, which is different from human conversations where there are many overlaps, interruptions, and silences~\citep{zimmerman19969}. 

Current human-LLM interactions require one participant to remain idle while the other generates responses. Interruptions are manually triggered with a ``stop'' button or certain keywords, resulting in conspicuously artificial communication. In human conversations, participants simultaneously process incoming information and formulate responses, often in overlapping and interleaved contexts, thus allowing each other to interrupt or be interrupted.

We introduce the concept of \textbf{duplex models} to address this limitation. 
Duplex models emulate human cognitive processes by synthesizing responses internally while simultaneously attending to incoming user inputs, akin to a person thinking while listening as well as speaking while observing. However, present autoregressive models face substantial challenges in adopting a duplex configuration, as they must process and encode a complete input message before generating any tokens, resulting in a turn-based conversation.

Considering this, we propose a framework for quickly converting current LLMs into duplex models by processing queries and responses pseudo-simultaneously without significant alternations to their architectures.

Specifically, we propose a time-division-multiplexing (TDM) encoding-decoding strategy. Messages in dialogues are split into time slices and the model processes time slices of input queries incrementally and generates time slices of output responses based on these partial input slices. 
When a new input query arrives, the model immediately halts its current generation process and starts a new sequence that integrates the additional input, enabling swift responses.
To adapt existing LLMs to this format of time slices, we build a duplex dataset for fine-tuning. The differences between our data from the conventional supervised fine-tuning (SFT) dataset are: (1) its input and output are time slices and (2) it includes various interactive user interruptions, such as generation termination, regeneration, and dialogue reset. 

To demonstrate the feasibility of duplex models, we train a prototype named MiniCPM-duplex, based on MiniCPM---a robust and lightweight LLM~\citep{hu2024minicpm}. 
Empirical results show that MiniCPM-duplex has its original performance on general benchmarks while enabling dynamic responses to user queries.
Additionally, we conduct a user study to compare the MiniCPM-duplex with the original MiniCPM. The results indicate that duplex models show significant improvements in responsiveness, human-likeness, and user satisfaction. Our contributions are fourfold:

    (1) We introduce and define the concept of duplex models, which are designed to generate output simultaneously as they receive input.

    (2) We propose a TDM encoding-decoding strategy and a duplex-specific SFT dataset for implementing duplex models.

    (3) We confirm that segmenting time slices during interactions does not compromise performance, and notably enhances the responsiveness, human-likeness, and overall satisfaction of conversations.

    (4) We release the model and dataset and provide a demo for users to experience firsthand.

\section{Duplex Models}

We define \textit{duplex models} as models that can process inputs and produce outputs simultaneously, and dynamically decide when to respond. It differs from current LLMs-based chatbots where participants must specify the end of inputs and only produce outputs after processing the entire input. To convert existing LLMs into duplex models, we split conversation messages into time slices, and then propose a TDM encoding-decoding mechanism to process these slices. 
To enhance the processing of these time slices, we further introduce duplex alignment to adapt existing LLMs to duplex models.

\begin{figure*}[t]
    \centering
    \subfloat[\label{fig:traditional_lm}]{\includegraphics[width=0.95\linewidth]{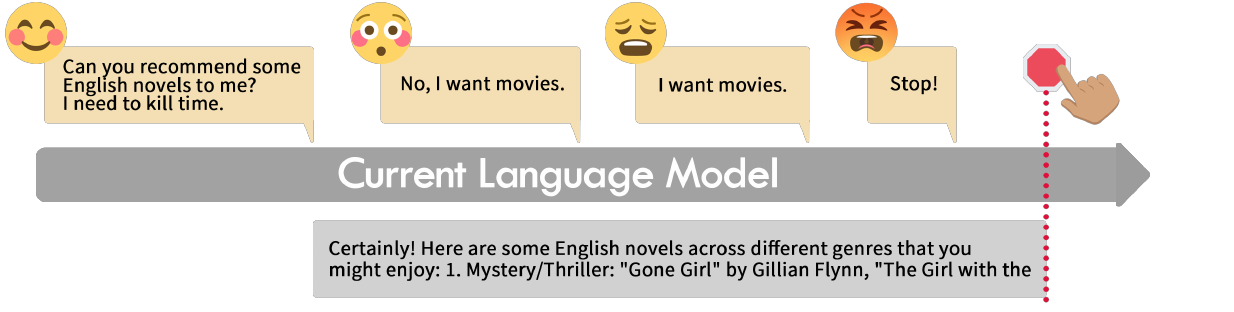}}\\
    \subfloat[\label{fig:duplex_lm}]{\includegraphics[width=0.95\linewidth]{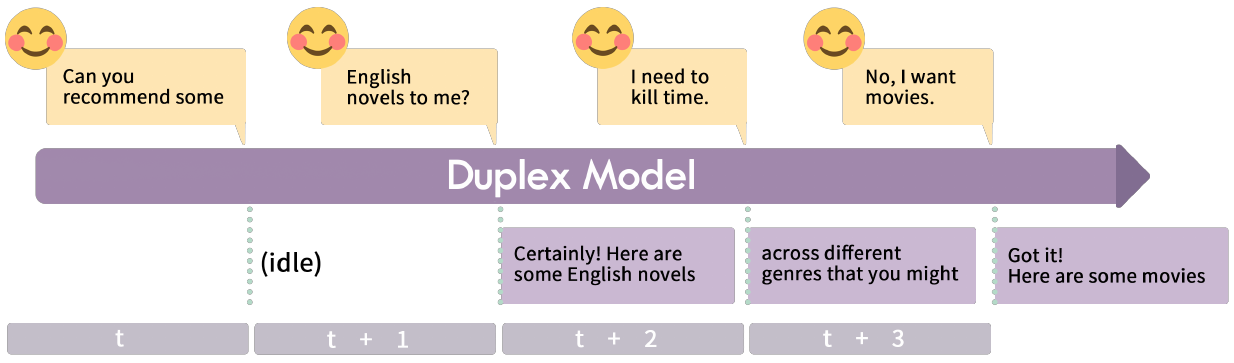}}
    \caption{Illustration of the input/output processing scheme of traditional models (\ref{fig:traditional_lm}) and duplex models (\ref{fig:duplex_lm}). Traditional models receive the complete input from the user before generating the response. In contrast, duplex models process the input and generate the output in an online manner. }
    \label{fig:overview}
\end{figure*}

\subsection{Time-Division-Multiplexing Encoding-Decoding} 

Current autoregressive language models struggle to function as true duplex systems. During the input phase, the LLM encodes the input into key-value caches without generating any output. To leverage autoregressive models in approximating duplex models, we propose a TDM strategy. We divide the conversation interaction into time slices and process input slices immediately to produce corresponding output slices. 

Instead of requiring users to specify when the model should respond, the duplex model infers responses after every 
$k$ seconds, i.e.,  each time slice spans $k$ seconds.
A special token (e.g., \texttt{<idle>}) is used to indicate the model's decision to remain silent and wait for further inputs. 
If not used, the generated slice is delivered to the user immediately. This approach mimics human conversational patterns more closely, as humans do not use special tokens to signal the end of utterances and intuitively determine the appropriate moments to respond to inputs.
Figure~\ref{fig:overview} illustrates the distinction between duplex and conventional language models.

\subsection{Time-Slicing Chunking}
\label{sec:chunk-sizes}

As shown in Figure~\ref{fig:overview}, all the input queries and output responses of conversations are in the slice format. 
The size of slices has great implications for the performance of a duplex model.
Large slice sizes result in greater response (or interruption) latency, while smaller slice sizes may result in unnecessarily long inputs (because some tokens are added between the chunks).
Our preliminary investigation and pilot experiments with our transformer-based~\citep{vaswani2017attention} models reveal that time-slicing chunking at 2-second intervals balances response latency and user experience. 
Assuming human beings usually speak 110-170 words per minute\footnote{\url{https://debatrix.com/en/speech-calculator/}}, an appropriate size of time slices is 4-6 words.

\begin{figure}
    \centering
    \resizebox{0.6\linewidth}{!}{
    \includegraphics{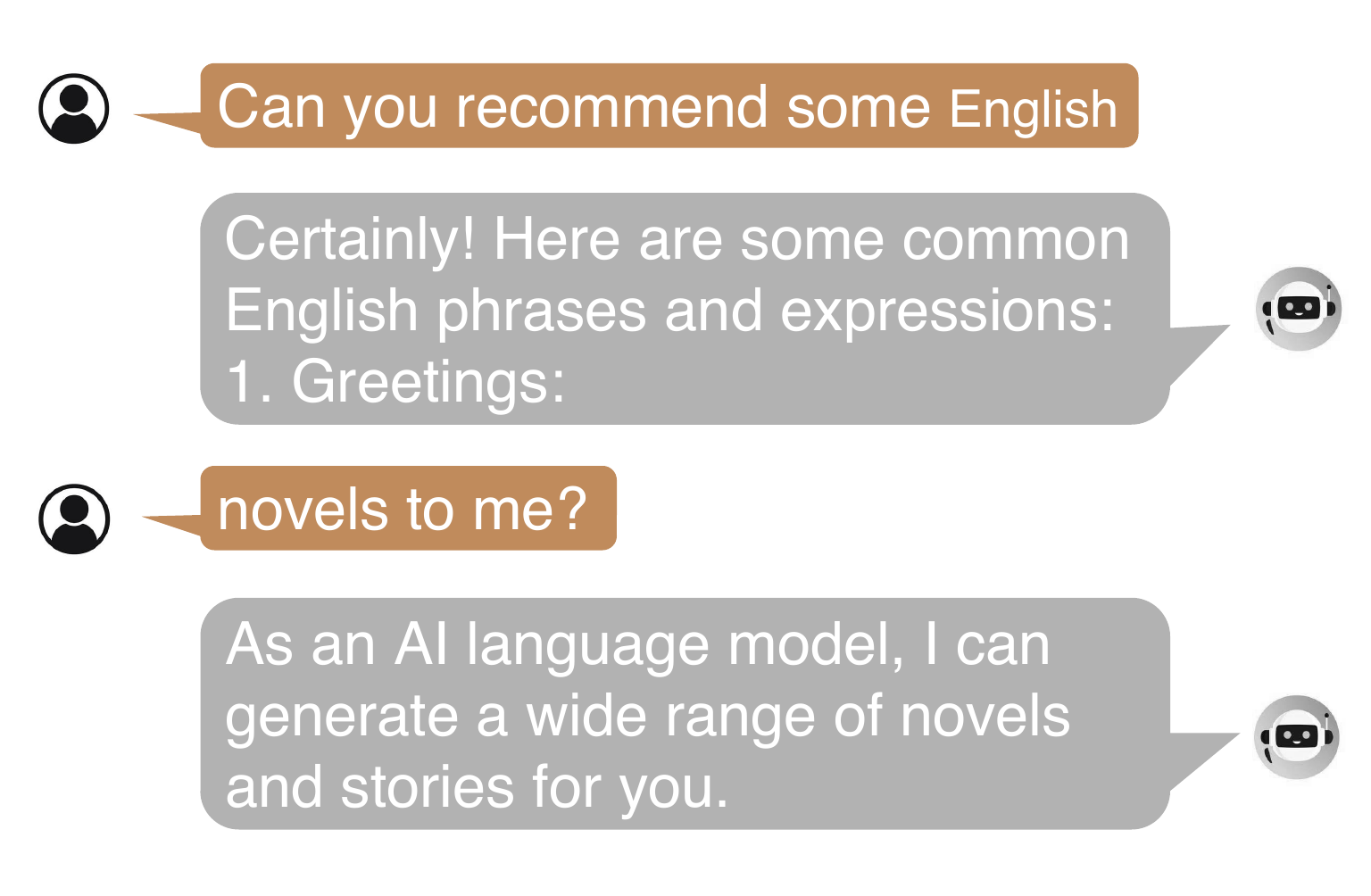}
    }
    \caption{Responses of MiniCPM when inputs are time slices.}
    \label{fig:normal_case}
\end{figure}
\subsection{Duplex Alignment}
Normal LLMs are unable to handle time slices as shown in Figure~\ref{fig:normal_case}, so we need to fine-tune them into duplex models. To achieve this, we construct a duplex SFT duplex dataset.

\section{Supervised Fine-Tuning Duplex Dataset}
\label{sec:data}

We create \textbf{Duplex-UltraChat} for tuning current LLMs into duplex models. Different from existing dialogue datasets, Duplex-UltraChat has no special tokens or keywords to indicate the beginning or end of messages. Messages are split into time slices. A slice is either the actual message of an individual or a special ``idle'' token to indicate silence. Each individual may interrupt by generating a response before the other party's message is completed.

Duplex-UltraChat is derived from UltraChat~\citep{ding2023enhancing} to reduce annotation costs. We heuristically inject appropriate random interruptions to simulate realistic scenarios. Powerful LLMs rewrite the interruptions to ensure diversity and naturalness.
Each user message is randomly split into 4-6 words. Assistant messages are split into 10-token slices.

During the construction of the dataset, we abide by the following two design choices: user behavior is unpredictable and the assistant should be polite.
Examples in the dataset can be categorized as uninterrupted dialogues and dialogues with interruptions.
As shown in Table~\ref{tab:duplexultrachat}, there are six categories of duplex data consisting of over 4.8M dialogues. Each piece of data has an average length of 2,570.2 tokens encoded by the tokenizer of MiniCPM-duplex and 170.4 slice pairs.

\begin{figure}
    \centering
        \subfloat[Basic]{\includegraphics[width=0.485\linewidth]{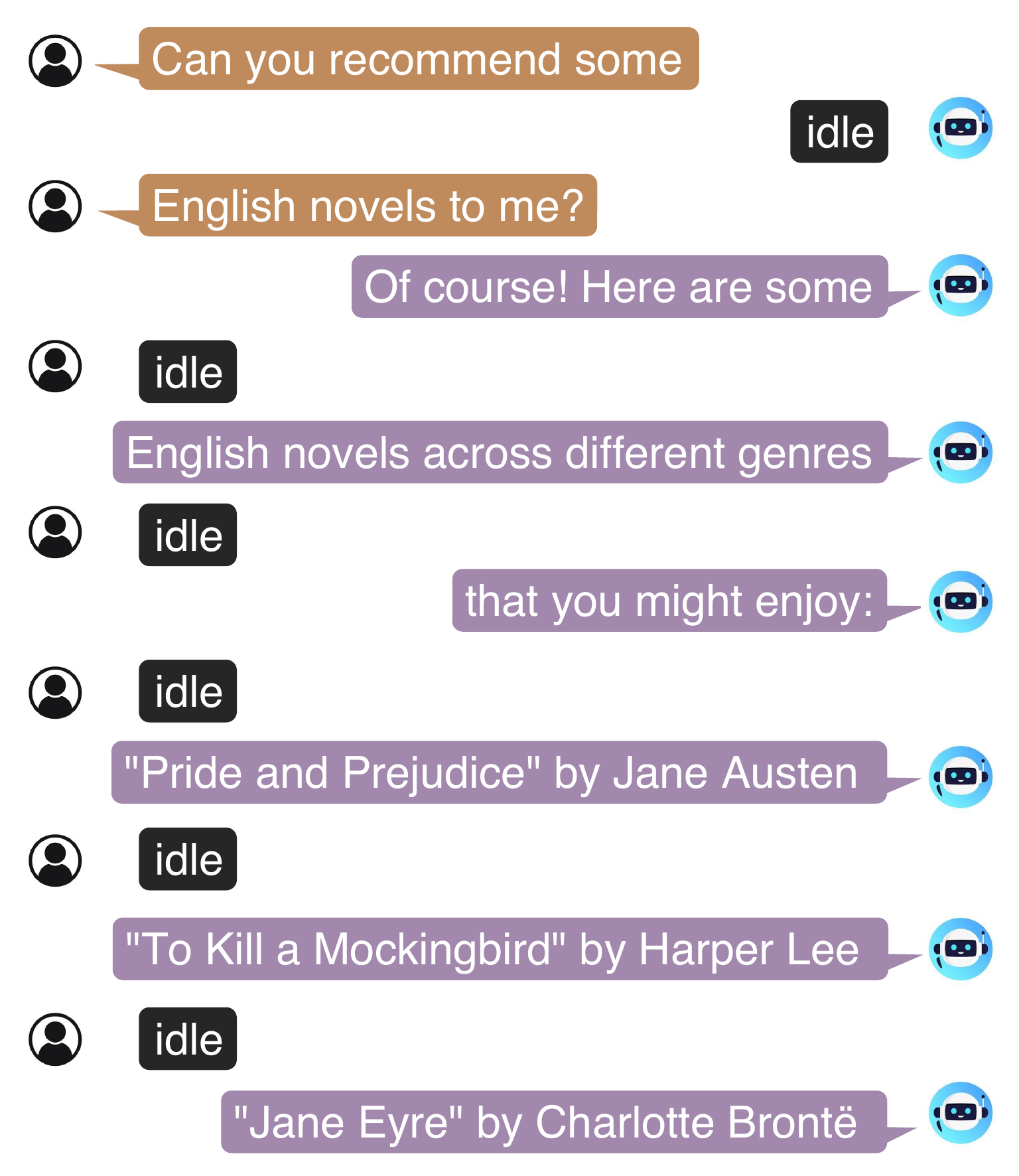}}
    \hspace{2pt}
            \subfloat[Topic interweaving]{\includegraphics[width=0.485\linewidth]{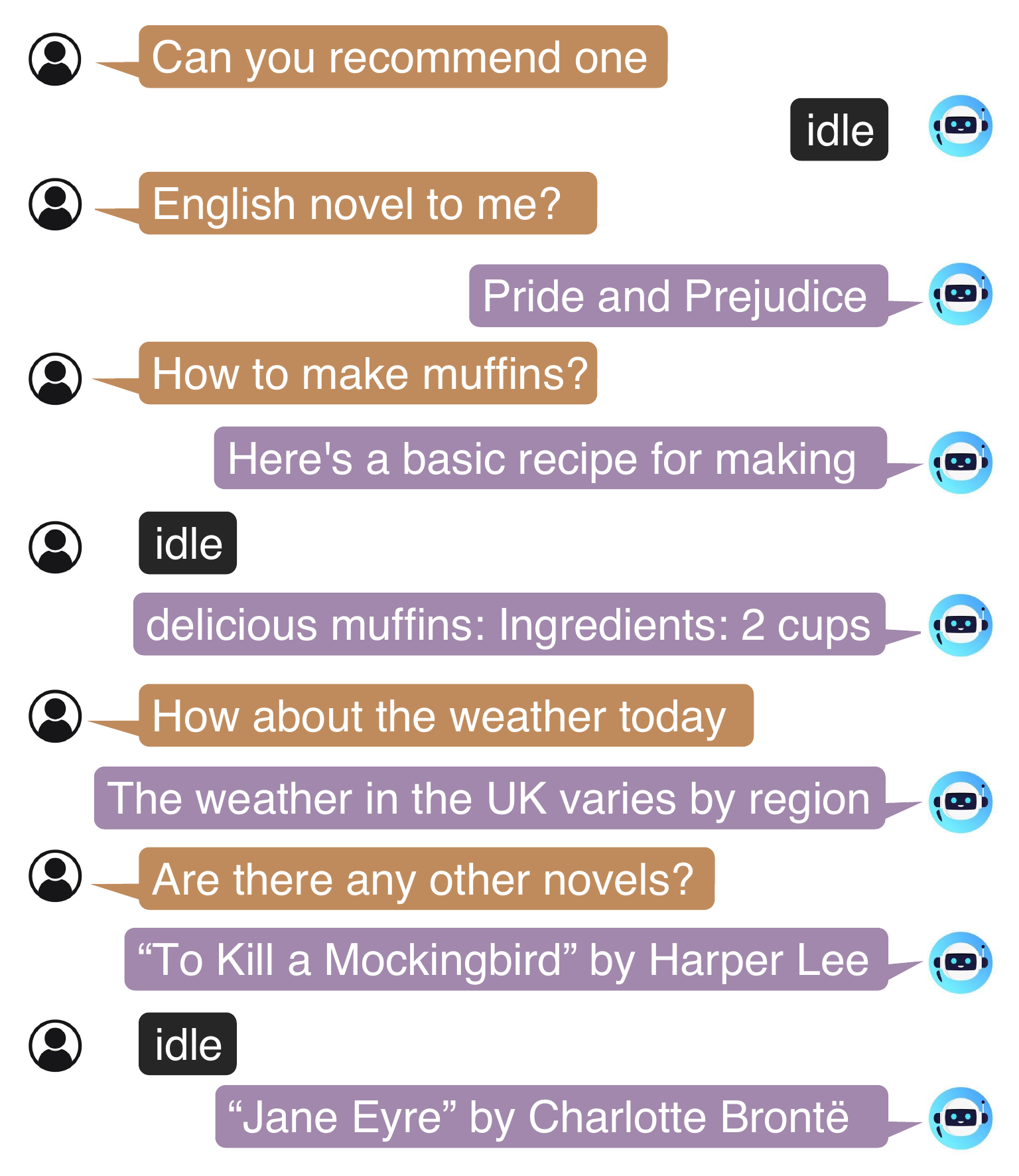}}

    \caption{An example of uninterrupted dialogue in Duplex-UltraChat.}
    \label{fig:basic}
\end{figure}
\begin{figure*}[!htp]
    \centering
    \subfloat[Termination]{\includegraphics[width=0.242\linewidth]{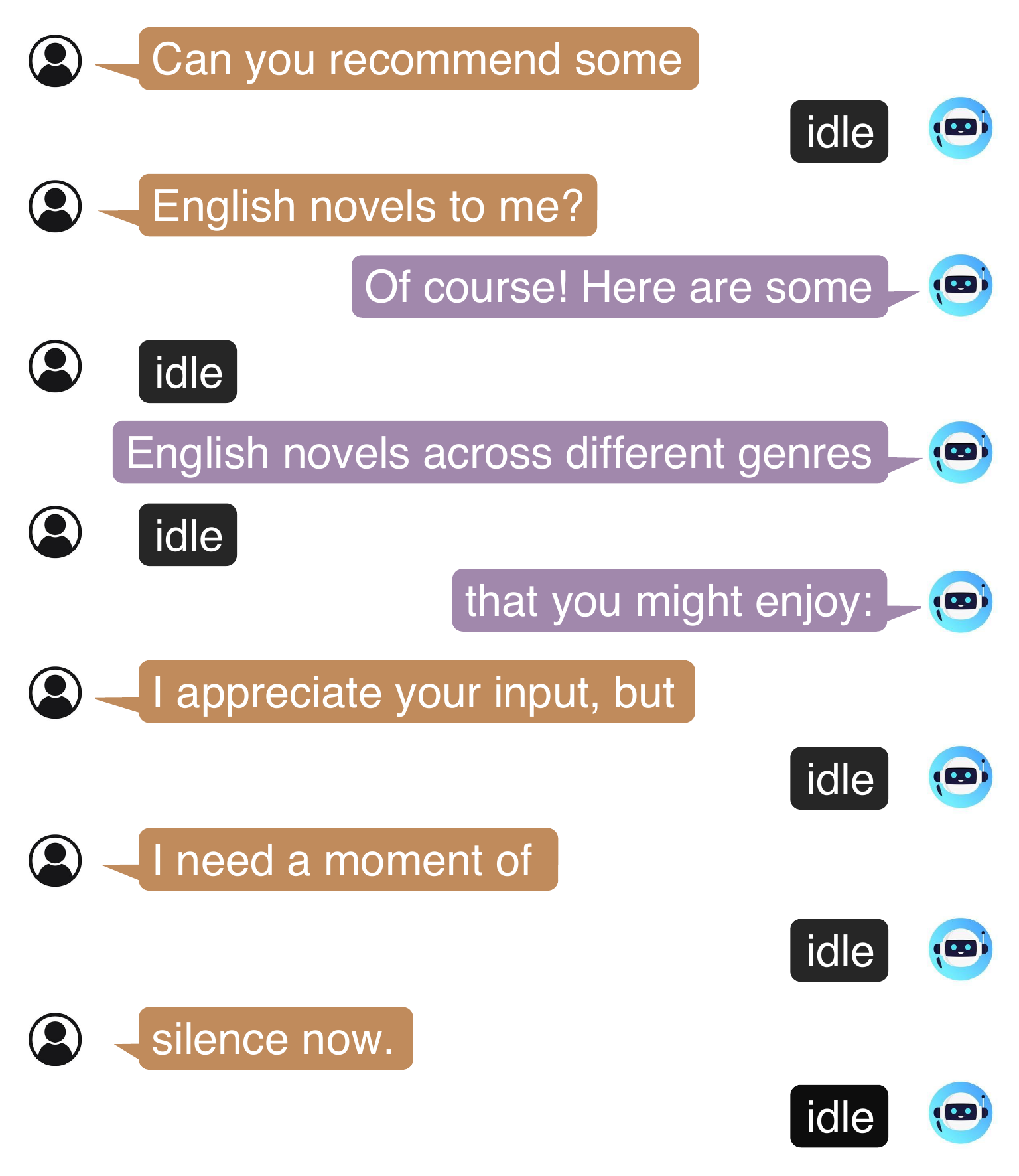}}
    \hspace{2pt}
    \subfloat[Regeneration]{\includegraphics[width=0.242\linewidth]{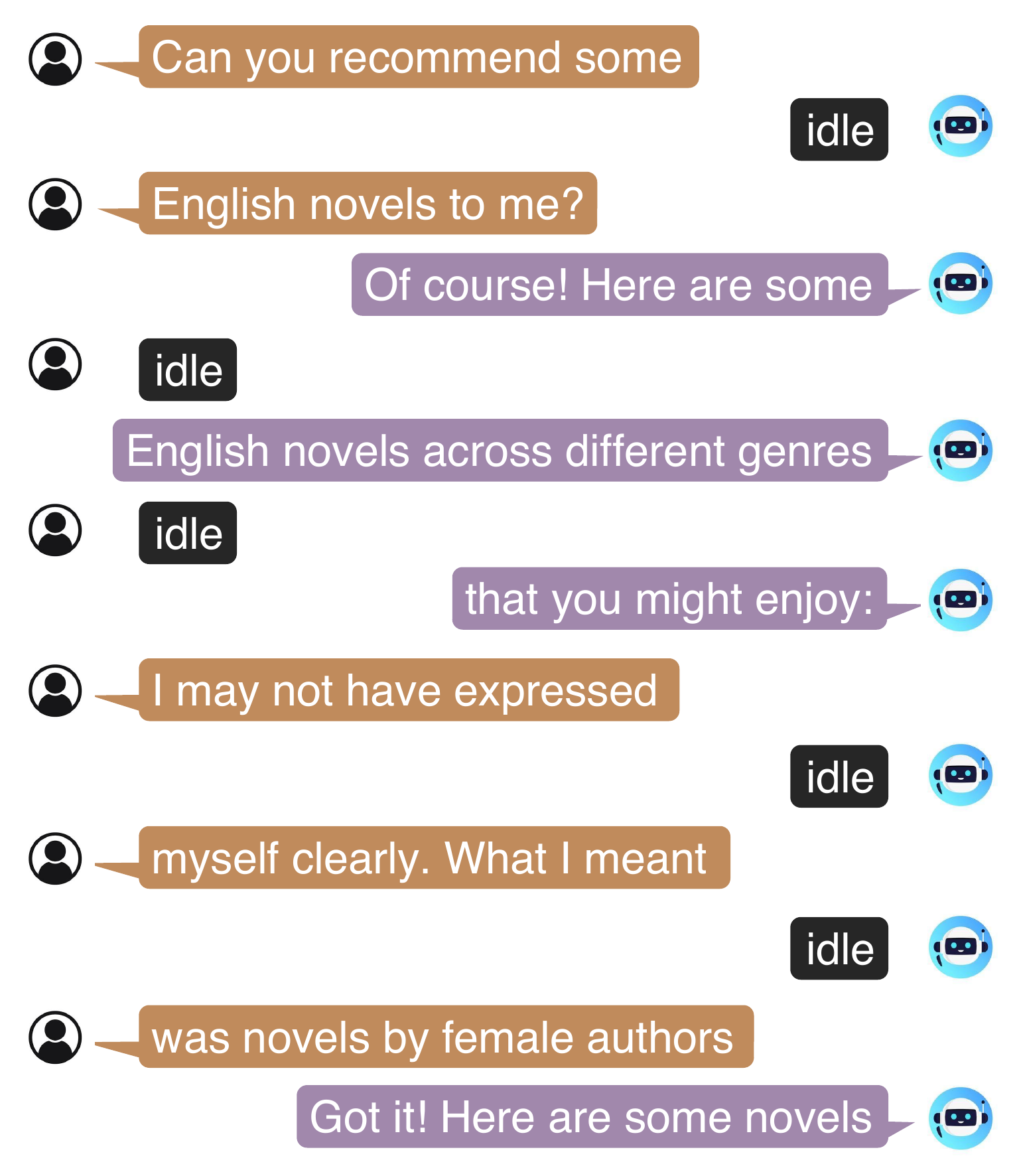}}
    \hspace{2pt}
    \subfloat[Dialogue reset]{\includegraphics[width=0.242\linewidth]{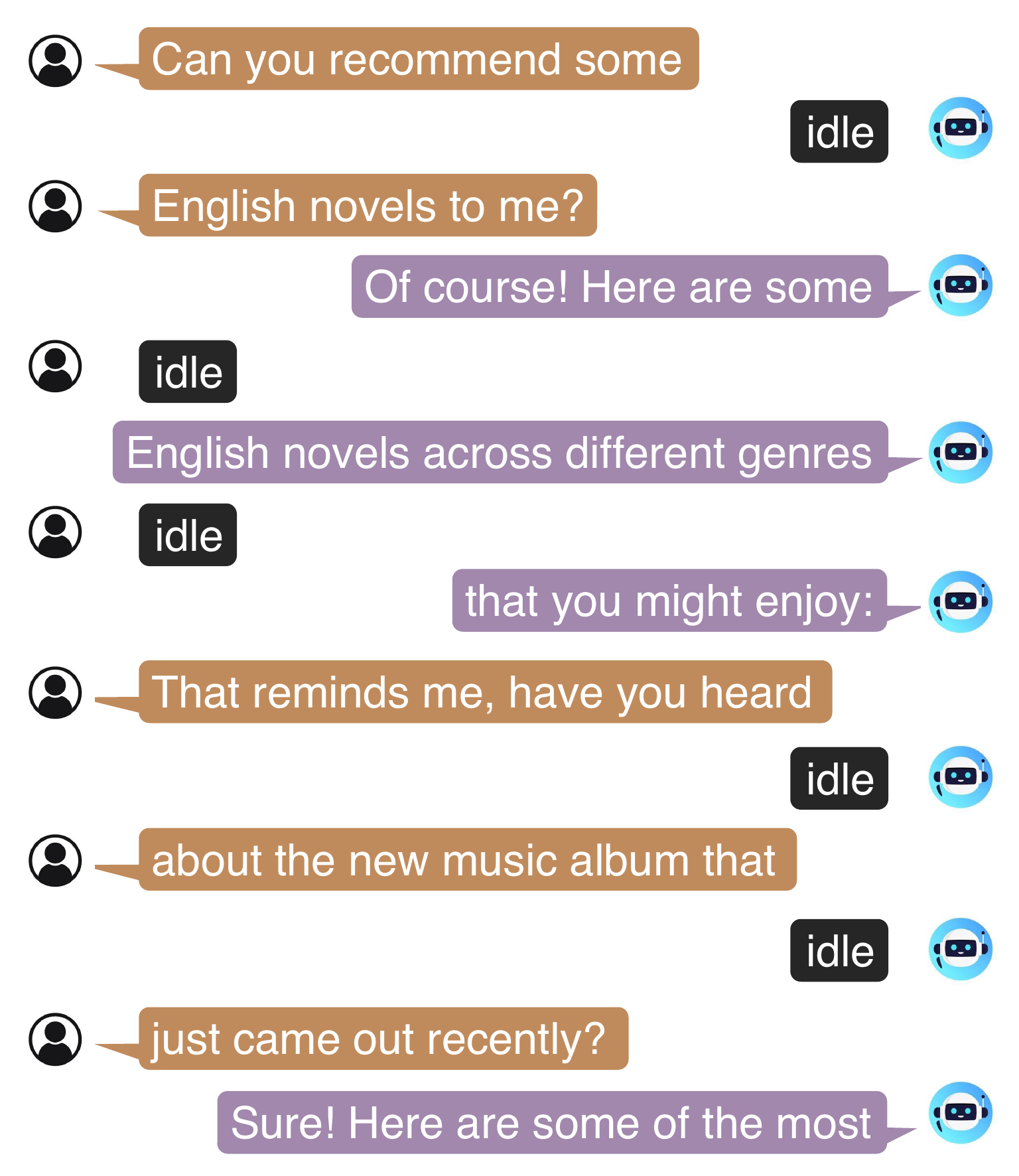}}
    \hspace{2pt}
    \subfloat[Back on topic ]{\includegraphics[width=0.242\linewidth]{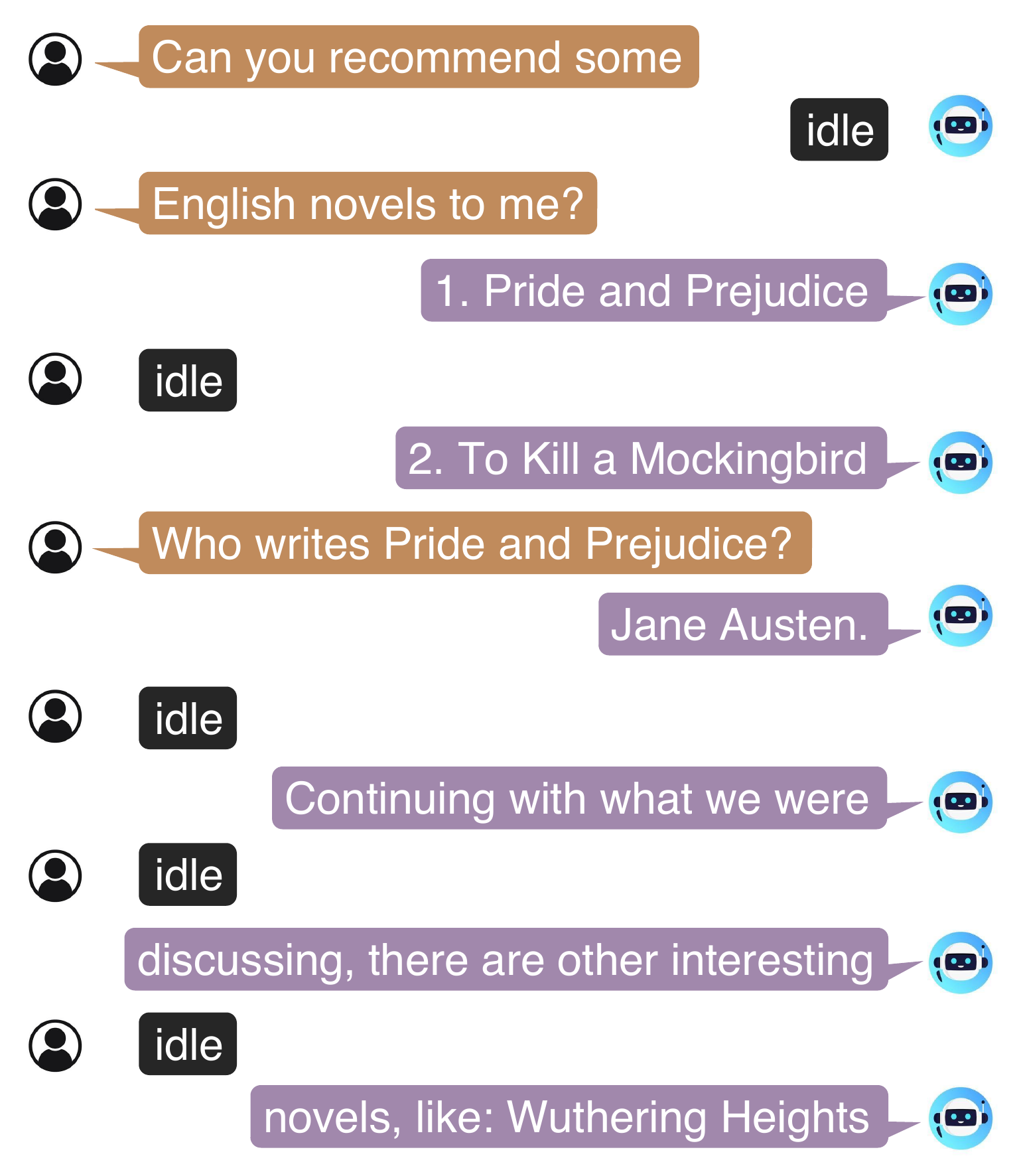}}
    \caption{Some examples from Duplex-UltraChat.}
    \label{fig:type1}
\end{figure*}

\begin{table*}[!t]
    \centering
    \resizebox{0.75\linewidth}{!}{
    \begin{tabular}{c|cccc}
    \toprule
         \textbf{Example Type} 
             & \textbf{Interruption}
         & \textbf{\# Dialogues} 
         & \textbf{Avg. \# Slice Pairs} 
         & \textbf{Avg. \# Tokens} \\
         \midrule
         Basic & \xmark & 1,458,353 & 153.9 & 2,342.2 \\
         Topic Interweaving & \xmark & 489,065 & 427.7 & 6819.6 \\
         Generation Termination & \cmark & 1,468,141 & 89.3 & 1,318.0\\
         Regeneration & \cmark & 806,687 & 171.2 & 2,590.4\\
         Dialogue Reset & \cmark & 300,318 & 194.7 & 2,906.5\\
         Back on Topic & \cmark & 327,286 & 199.1 & 2495.6\\
         \midrule
         Total &  & 4,849,850 & 170.4 & 2,570.2 \\
         \bottomrule
    \end{tabular}
    }
    \caption{The statistics of Duplex-UltraChat. The tokenizer of our MiniCPM-duplex produces the tokens.}
    \label{tab:duplexultrachat}
\end{table*}

\subsection{Uninterrupted Dialogue}

\paragraph{Basic} 

Ordinary uninterrupted dialogue data is obtained by splitting existing dialogue messages into slices. When the user input is unfinished, the output of the duplex model should be \texttt{<idle>}. Meanwhile, when the duplex model is generating output, the user is set to quiet and its input is \texttt{<idle>}. Figure~\ref{fig:basic} shows an example of basic duplex data.

\paragraph{Topic Interweaving} 

People may discuss several topics interweavingly ignoring coherence. To mimic such behavior, we interlace sentences of 3-5 dialogues while keeping their orders, and split each sentence into time slices as the basic type does.

\subsection{Dialogues with Interruptions}

In realistic human conversions, the individuals may start speaking before the other part is done with their message. Therefore, to simulate such scenarios, we inject four interruptions into the data as shown in Figure~\ref{fig:type1}.

\paragraph{Generation Termination} 

Forced interruptions are when users directly speak out their next sentence regardless of the status of the assistant. 
To generate such data, we randomly choose a location in an assistant message, discard the remaining part of the message, and insert a new user input at that location. We prefix the user input with one of the 11 pre-defined transitional sentences (see Appendix~\ref{sec:interruption_transition}). ChatGPT rewrites this input to ensure a natural and varied transition. The target output is idle tokens because the assistant is expected to terminate its current response.

Generation termination requires the assistant to learn to stop speaking when the user is forcibly interrupting it and be robust to incomplete messages in the chat history. Since this interruption may be regarded as impolite, our dataset does not contain situations where the user is interrupted.

\paragraph{Regeneration} 

Another scenario where the user interrupts the assistant is when the user is dissatisfied with the current response. 
In conventional LLM-based chatbots, the user must first stop the generation with a button, and prompt the model with the updated prompt. In contrast, duplex models allow the user to directly interrupt and reinput the new prompt while generating outputs.
To create such data, we randomly sample a user message and repeat it with one of 15 pre-defined transition sentences (given in Appendix \ref{sec:mistake_transition}). ChatGPT rewrites this repetition message for better coherence. Then, the chat history and repetition message are fed to ChatGPT to generate the annotation.

\paragraph{Dialogue Reset} 

Here, we consider situations where the user wants to chat abruptly on an entirely different topic while the assistant is generating output.
To create such data, we randomly sample five dialogues and truncate the first four dialogues at random locations before concatenation.
We define 18 kinds of transitional sentences in Appendix~\ref{sec:topic_transition}, including one empty string. We randomly choose a transitional sentence, and prefix it with the first sentence of the new dialogue. Each message is then rewritten by ChatGPT. If the selected transitional sentence is the empty string, we do not rewrite the input, which simulates certain users who wish to start a new dialogue as fast as possible.

\paragraph{Back on Topic} 

When the user only interrupts a question without attempting to stop the assistant or change the topic, the assistant should answer the question and then continue the unfinished statement. 
To construct this type of data, we randomly select a within a message from the assistant, and annotate a question about a statement by the assistant. 
GPT-4~\citep{gpt4} is used to generate the answer to the user's question and continue the interrupted message with coherence.

\section{Experimental Details}

\begin{figure*}[!t]
    \centering
    \subfloat[]{\includegraphics[width=0.15\textwidth]{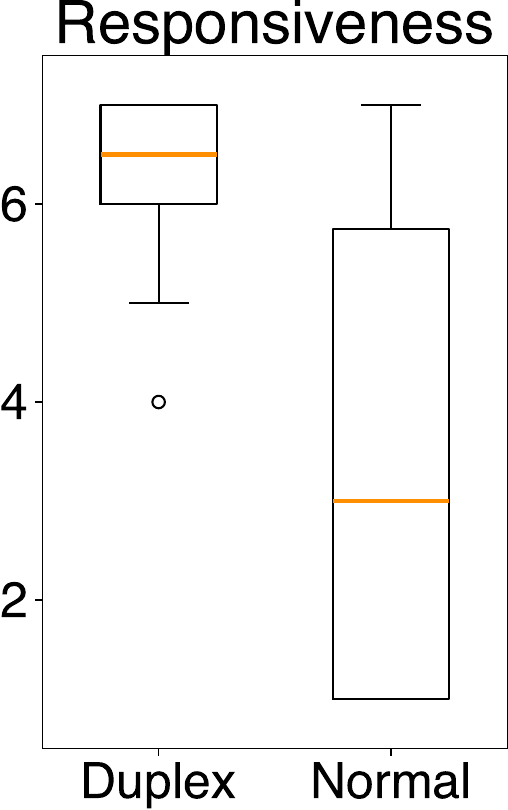}}
    \hspace{5pt}
    \subfloat[]{\includegraphics[width=0.15\textwidth]{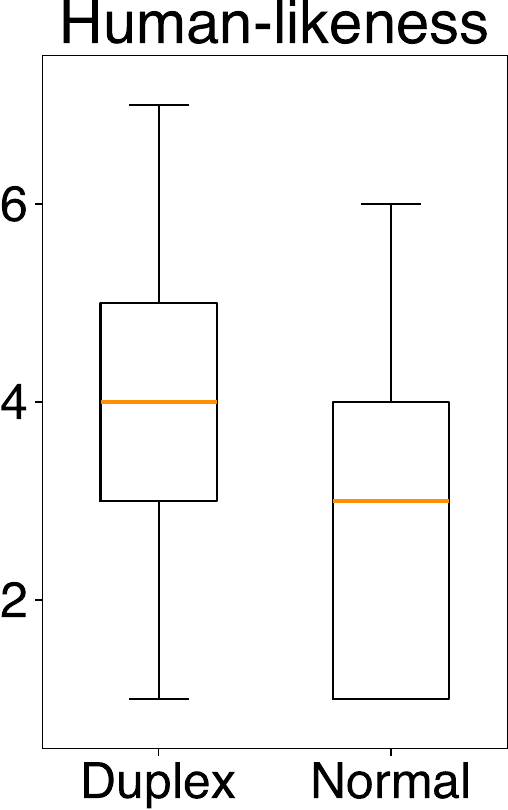}}
    \hspace{5pt}
    \subfloat[]{\includegraphics[width=0.15\textwidth]{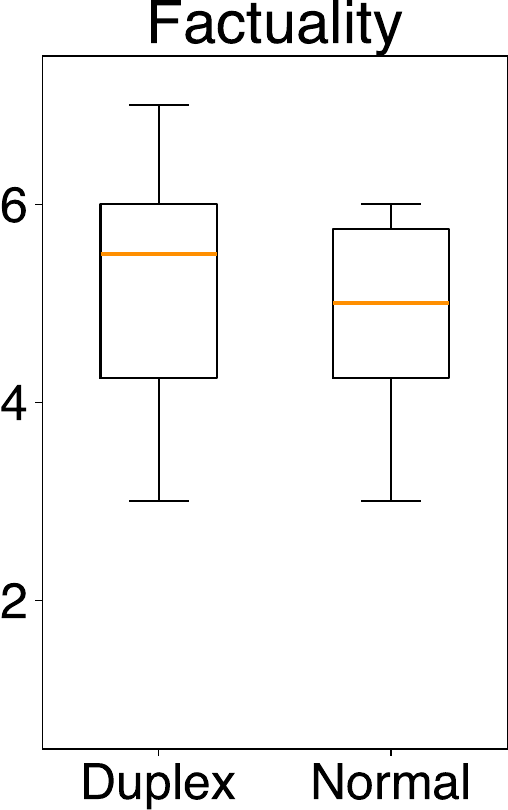}}
    \hspace{5pt}
    \subfloat[]{\includegraphics[width=0.15\textwidth]{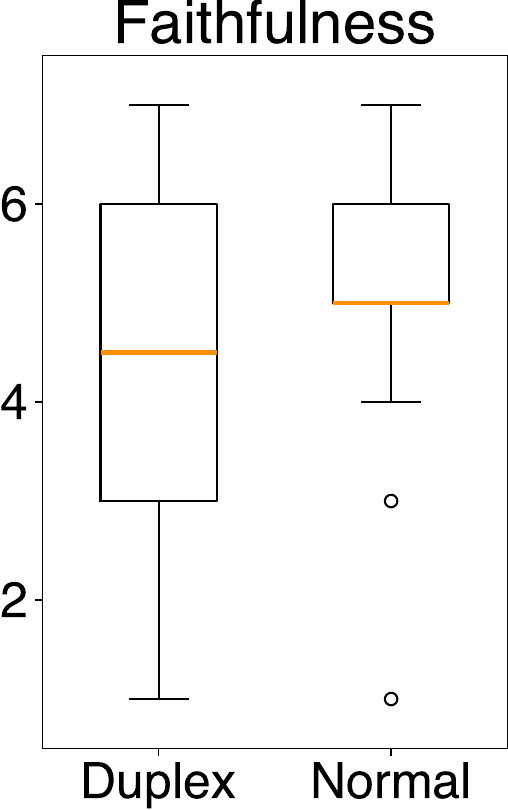}}
    \hspace{5pt}
    \subfloat[]{\includegraphics[width=0.15\textwidth]{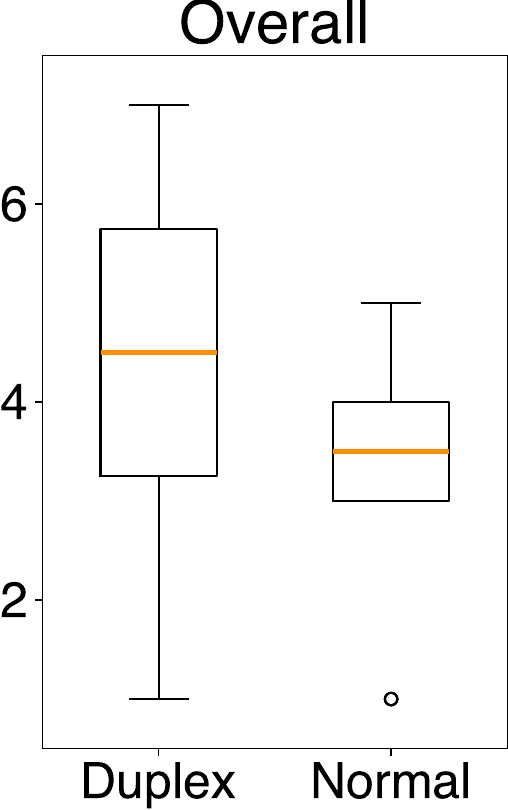}}
    \caption{The human evaluation score distributions for MiniCPM and MiniCPM-duplex regarding responsiveness, human-likeness, factuality, faithfulness, and overall satisfaction.}
    \label{fig:human-evaluation-box}
\end{figure*}
\subsection{Training}


We start from the public checkpoint of MiniCPM-2.4B~\citep{hu2024minicpm}\footnote{\url{https://huggingface.co/openbmb/MiniCPM-2B-sft-bf16}, denoted MiniCPM.} and fine-tune it on Duplex-UltraChat as well as the SFT data that MiniCPM uses to obtain MiniCPM-duplex.

We make the following modifications to MiniCPM: (1) we append a special end-of-sentence token (i.e., \texttt{<eos>}) to each response of the duplex model, and (2) we add a special token \texttt{<idle>} to represent empty input or output.

The training of MiniCPM-duplex uses the following hyperparameters: $10^{-3}$ maximum learning rate, Warmup-Stable-Decay~\citep{hu2024minicpm} learning rate scheduler, a batch size of 800, and a maximum length of 4,096. 
We train for 10,000 steps on 40 NVIDIA A100 GPUs for 36 hours.

\subsection{Baseline}

Since our MiniCPM-duplex and MiniCPM are derived from the same checkpoint, we verify the effectiveness of our method by comparing it against the vanilla MiniCPM.

\subsection{Evaluation}

We evaluate the duplex model with three kinds of metrics: automatic metrics, GPT-4, and human. Automatic metrics, like accuracy and pass rate, are widely used for convenience and low cost. 

\subsubsection{GPT-4 Evaluation}

To evaluate the multi-turn dialogue ability of MiniCPM-duplex, we benchmark it on 
MT-Bench~\citep{zheng2024judging} and MT-Eval\citep{kwan2024mt} with GPT-4 as the judge.

To mimic real-time scenarios, we chunk each instruction in MT-Bench and MT-Eval into multiple 4-6 word slices and feed one slice at a time. Then we concatenate all output segments from the duplex model to form the final output. For the traditional model, we directly feed the entire prompt to the model.

Both models use the same decoding parameters: random sampling, a temperature of 0.8, a top-$p$ value of 0.8, and a top-$k$ value of 0. The maximum length is set to 4,096. For the duplex model, we set the maximum token generated per chunk to 10.

\begin{figure}[!t]
    \centering
    \includegraphics[width=0.98\linewidth]{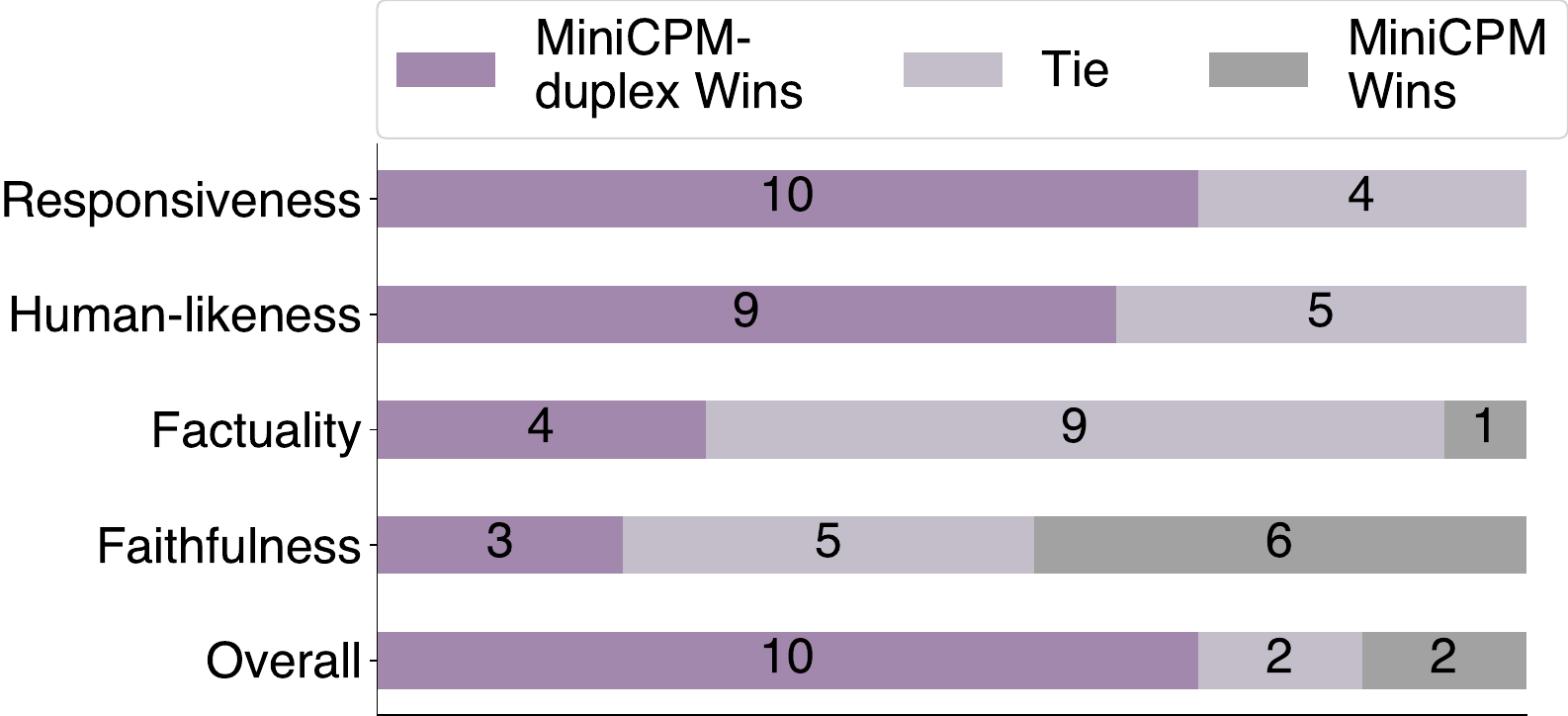}
    \caption{Win rates between MiniCPM and MiniCPM-duplex on responsiveness, human-likeness, factuality, faithfulness, and overall satisfaction.}
    \label{fig:human-evaluation}
\end{figure}

\subsubsection{Human Evaluation}

When using humans as evaluators, we consider the following four aspects.

\paragraph{Responsiveness} 

This metric measures whether a model will respond to a user query and the latency if it responds, which is a perceived latency. 
Many factors may contribute to greater response latency, including the speech-to-text and text-to-speech conversion time, model inference time, network latency, and the interaction strategy that the model utilizes. There is no obvious difference between the actual inference latency of MiniCPM-duplex and MiniCPM.

\paragraph{Human-Likeness} 

Inspired by the Turing test, we wish to develop a language model that chats in a way indistinguishable from humans. 
Therefore, we define human-likeness as a metric that measures the degree of the similarity of a model to humans.

\paragraph{Faithfulness} 

Faithfulness is a widely used metric in the evaluation of LLMs~\citep{arras2017relevant,serrano-smith-2019-attention,jain-wallace-2019-attention,deyoung2020eraser,adlakha2023evaluating,chen2023improving}. Here, we use it to reflect the degree how the model follows a user's instruction, which is similar to~\citep{adlakha2023evaluating}. 

\paragraph{Factuality} 

This metric measures the degree of hallucination of a LLM~\citep{rudinger2018neural,tian2023fine,chen2023beyond,wang2023survey,nakano2021webgpt}.

\subsection{Interactive Demo} 

We implement an interactive demo with a user interface such that human evaluators can make evaluations based on actual interaction experience.
In the demo, users chat with an assistant using voice. The assistant is either implemented with the vanilla MiniCPM or our MiniCPM-duplex. The conversion between speech and text is implemented with Google's cloud-based ASR and TTS API\footnote{Speech-to-text API: \url{https://cloud.google.com/speech-to-text/docs/reference/rest}. Text-to-speech API: \url{https://cloud.google.com/text-to-speech/docs/reference/rest}.}.

This demo supports both vanilla MiniCPM and MiniCPM-duplex. 
For the vanilla MiniCPM, the program automatically detects pauses in the user's voice. On each pause, the speech is converted to text, which is then sent to the model. MiniCPM performs regular text generation, and each output token is passed to the ASR module, before being returned to the user. Meanwhile, the user has to wait until the speech response is done before the next query. 
When interacting with MiniCPM-duplex, the user's speech is processed every 2 seconds.
When the MiniCPM-duplex does not generate the idle token, the text generation will be transcribed into audio and played out. The user's voice will be captured, transcribed, and fed to the model regardless of whether the assistant speaks.

\begin{table}[!ht]
    \centering
    \resizebox{0.9\linewidth}{!}{
    \begin{tabular}{l|cc}
        \toprule
        \textbf{Benchmark} & \textbf{MiniCPM} & \textbf{MiniCPM-duplex}\\
        \midrule
        C-Eval  & \textbf{50.52}& 50.06\\
        CMMLU  &\textbf{51.30} & 48.53\\
        MMLU & 53.45& \textbf{53.76}\\
        BBH & \textbf{37.25}& 36.35\\
        \midrule
        HumanEval &\textbf{50.00} & 49.39\\
        MBPP & 38.09&\textbf{38.30} \\
        \midrule
        GSM8K &42.30 & \textbf{46.10}\\
        MATH & \textbf{10.56}&9.32 \\
        \midrule
        ARC-e & 84.60&\textbf{85.19} \\
        ARC-c & 69.80& \textbf{70.05}\\
        HellaSwag &\textbf{61.40} & 60.79\\
        \bottomrule
    \end{tabular}
    }
    \caption{Performances of MiniCPM and MiniCPM-duplex on standard benchmarks.}
    \label{tab:commonbench}
\end{table}
\begin{table}[!ht]
    \centering
    \resizebox{0.9\linewidth}{!}{
    \begin{tabular}{l|cc}
        \toprule
        \textbf{Metric} & \textbf{MiniCPM} & \textbf{MiniCPM-duplex}\\
        \midrule
        Responsiveness & 3.43 & \textbf{6.21}\\
        Human-Likeness & 2.79 & \textbf{4.00}\\
        Factuality & 4.93 & \textbf{5.21}\\
        Faithfulness & \textbf{5.14} &4.50\\ 
        Overall & 3.29 & \textbf{4.36}\\
        \bottomrule
    \end{tabular}
    }
    \caption{Average human evaluation scores on responsiveness, human-likeness, factuality, faithfulness, and overall satisfaction. Higher is better.}
    \label{tab:human-evaluation}
\end{table}

\begin{table}[!ht]
    \centering
    \resizebox{0.9\linewidth}{!}{
    \begin{tabular}{l|cc}
        \toprule
        \textbf{Score} & \textbf{MiniCPM} & \textbf{MiniCPM-duplex}\\
        \midrule
        First turn & \textbf{7.17} & 5.83 \\
        Second turn& \textbf{5.85} & 4.84 \\
        \midrule
        Avg.& \textbf{6.51} & 5.33 \\
        \bottomrule
    \end{tabular}
    }
    \caption{MT-Bench results of MiniCPM and MiniCPM-duplex. Higher is better.}
    \label{tab:mtbench}
\end{table}

\begin{table}[!ht]
    \centering
    \resizebox{0.9\linewidth}{!}{
    \begin{tabular}{l|cc}
        \toprule
        \textbf{Score} & \textbf{MiniCPM} & \textbf{MiniCPM-duplex}\\
        \midrule
        Refinement-multi & \textbf{5.87} & 5.78 \\
        Expansion-multi& \textbf{6.31} & 5.80 \\
        Follow-up-multi& 8.48 & \textbf{8.56} \\
        \midrule
        Avg.& \textbf{6.88} & 6.71 \\
        \bottomrule
    \end{tabular}
    }
    \caption{MT-Eval results of MiniCPM and MiniCPM-duplex. Higher is better.}
    \label{tab:mteval}
\end{table}

\subsection{User Study}

Specifically, we recruit 14 participants consisting of 5 males and 9 females from 18 to 35 years old. Each participant holds a Bachelor's or Master's degree. 
Details on employment, payment, and ethical review are in Appendix~\ref{app_human}.

During the experiment, we rename MiniCPM-duplex as Model A, and MiniCPM as Model B to ensure anonymity. Participants are unaware of the difference between the two models beforehand. We specify the odd-numbered participants interact with Model A first, and the even-numbered ones first chat with Model B to eliminate the influence of chatting order. When finishing chatting with a model, the participant should score it and continue interacting with the other one. After the experiment, participants could modify and confirm scores for both models. Each participant is assigned at least 5 sessions of multi-turn dialogues with each model. The first sentence of sessions should be the same for both models. To help the participants come up with topics to chat about, we provide them with a reference note containing sample instructions from AlpacaEval~\citep{alpaca_eval}.

\paragraph{Questionnaire Design} 

The questionnaire consists of six questions. The first five questions prompt the user to rate the model based on responsiveness, human-likeness, faithfulness, factuality, and overall experience.
The answer choices for these questions are scores from 1 to 7, where 1 represents disappointment, 4 represents indifference, and 7 represents excellence. 
The final question is open to suggestions on improving our duplex model. The actual questions are listed in Appendix~\ref{sec:appendix-survey-questions}.

\section{Results}
\label{sec:results}

\paragraph{Standard Benchmarks}

MiniCPM-duplex is benchmarked on several standard benchmarks, including multitask (C-Eval~\citep{huang2024c}, CMMLU~\citep{li2023cmmlu}, MMLU~\citep{hendrycks2020measuring}, BBH~\citep{suzgun2023challenging}), code (HumanEval~\citep{chen2021evaluating}, MBPP~\citep{austin2021program}), math (GSM8K~\citep{cobbe2021training}, MATH~\citep{hendrycks2021measuring}), and reasoning (ARC-e, ARC-c~\citep{clark2018think}, HellaSwag~\citep{zellers2019hellaswag}) with the LLM evaluation platform, UltraEval~\citep{he2024ultraeval}. Table~\ref{tab:commonbench} indicates that adapting to duplex models does not significantly harm its performance on general benchmarks.

\paragraph{GPT-4 Evaluation}

Table~\ref{tab:mtbench} and Table~\ref{tab:mteval} show the GPT-4 evaluation results on MT-Bench and MT-Eval, respectively. MiniCPM-duplex is slightly inferior to MiniCPM mainly due to that MiniCPM-duplex tends to generate shorter responses. GPT-4 favors longer responses, whereas users prefer chat models that give concise answers.

\paragraph{Human Evaluation}

We have received 14 questionnaire. Table~\ref{tab:human-evaluation} lists the average scores of both models on five metrics. The duplex model surpasses the normal model by 81.05\%, 43.37\%, and 32.52\% on responsiveness, human-likeness, and overall experience respectively. 

Apart from absolute scores, we compare the ratings of the two models and count the number of evaluators that rate one model higher.
The comparison results are shown in Figure~\ref{fig:human-evaluation}. 
MiniCPM is more faithful than the duplex model mainly because it uses more diverse SFT data. Whereas the duplex model wins in other aspects, with an exceptionally large margin on responsiveness and human-likeness. 

From these results, we conclude that duplex models can provide a better user experience in acting as the backbone model in AI assistants compared to ordinary language models.

\section{Analysis \& Discussion}

\subsection{Analysis}
\label{sec:analysis}

The superior performance of the duplex model is mainly due to its underlying receive/generate mechanism. 
Rather than strictly turn-based dialogue where users must explicitly signal the beginning and end of messages, duplex models behave more like human beings.
Besides, the duplex model has learned when to speak at the fine-tuning stage on the Duplex-UltraChat, which makes it more human-like. 
Such ability is essential in passing a non-turn-based version of the Turing test, which is a more realistic test for whether a machine can be indistinguishable from humans~\citep{barnaud2017perceptuo}.

\subsection{Discussions}

We highlight some important open problems associated with duplex models below.

\paragraph{High-quality duplex data is urgently needed}

Existing dialogue datasets are inherently turn-based, which does not represent realistic and complex human conversations. Despite some success in empirical results with our synthetically generated duplex dataset, it still lags behind the practical demands.
Two out of the 14 participants pointed out that they preferred concise responses rather than tedious answers. 

We manually inspect 10 out of 90 chat sessions and find that the duplex model fails to remain silent once and interrupts the user unexpectedly once, showing that there is room for improvement. 
Thus, high-quality duplex datasets are in urgent need. 

\paragraph{A new decoding strategy is needed to improve the chat experience} 

There are failed cases where the duplex model interrupted users unexpectedly.  
Balancing response speed and user experience is an open problem. 
Besides, to be more human-like, the duplex model should learn to start dialogues or topics actively. 

\paragraph{A custom TTS system is needed to smooth the output voice} 

The duplex model generates output chunk by chunk, which causes the output voice to be chunked. This results in incoherent intonation and volume, harming the user experience because existing TTS software does not support transcribing sequentially provided text chunks into a contiguous smooth voice. Overcoming this problem will improve the user experience considerably.

\section{Related Work}






\subsection{Dialogue Dataset}

Dialogue data can be divided into two categories: single-turn and multi-turn.

\paragraph{Single-Turn} 

Self-instruct~\citep{wang2023self} is a synthetic instruction-following dataset of over 82K instances generated by GPT-3.5. \citet{alpaca} adopt the data construction pipeline from~\citet{wang2023self} and construct Alpaca, a dataset with 52K instances. GPT-4-LLM~\citep{peng2023instruction} improves the Alpaca by replacing the data generator with GPT-4. It also adopts a Chinese version of Alpaca and Unnatural Instructions~\citep{honovich2023unnatural}. Besides, there are several high-quality datasets, such as BELLE~\citep{ji2023exploring} and GPT-4ALL~\citep{anand2023gpt4all}, among others.

\paragraph{Multi-Turn} 

DailyDialog~\citep{li-etal-2017-dailydialog} consists of over 13K dialogues annotated by humans, covering diverse daily conversation scenarios. Baize~\citep{xu-etal-2023-baize} generates multi-turn dialogues with ChatGPT by a prompting framework called self-chat where seed questions are from Quora and Stack Overflow, two popular question-answering websites. SODA~\citep{Kim2022SODAMD} contains dialogues involving social commonsense. UltraChat~\citep{ding2023enhancing} focuses on 30 meta-concepts and 20 types of materials and consists of over 1.4M dialogues.

\subsection{Dialogue Models}


Chat-based models have gained widespread popularity since the release of ChatGPT. Some notable chat-based LLMs include the Claude series~\citep{claude2.1,claude3}, Qwen series~\citep{Qwen1.5}, the Mistral series~\citep{jiang2023mistral} and LLaMa series~\citep{touvron2023llama}, among others. 
Most of these models, especially open-sourced ones, are purely text-based.

To enhance user experience, several applications support voice interaction.
One instance is ChatGPT, where users press a button before speaking and indicate the end of speech with a button or pausing \citep{chatgpt_assistant}.
Then ChatGPT processes the received signal and produces a response until it finishes or users interrupt it by pressing a button. 
Such an implementation is unrealistic because it requires the user to specify the beginning and end of inputs.
Whereas, our MiniCPM-duplex may improve this interactive experience by teaching the model to learn when to speak and when to be silent.

\section{Conclusion}

We have introduced the concept of duplex models and provided one implementation. To this end, we also constructed the first non-turn-based dialogue dataset, Duplex-UltraChat, by injecting diverse kinds of interruptions into existing dialogue datasets. Our model, MiniCPM-duplex, is competitive with traditional models when evaluated on ordinary benchmarks while outperforming them in terms of responsiveness, human-likeness,  and overall satisfaction. We believe that this work represents an essential step toward building machines that behave more human-like beyond current turn-based conversations.


\newpage

\section*{Limitations}

In this paper, we propose and verify the viability of duplex models. However, our implementation, MiniCPM-duplex, is a pseudo-duplex model, since it cannot perform encoding and decoding simultaneously. 
Consequently, our fixed-interval decoding strategy introduces a new hyperparameter that compromises responsiveness and context length (as discussed in Section \ref{sec:chunk-sizes}). These limitations call for a new architecture that better supports the input-output scheme of duplex models.

\section*{Acknowledge}

This work was supported by the National Key R\&D
Program of China (No.2022ZD0116312), Quan
Cheng Laboratory (Grant No.QCLZD202301), and the National Natural Science Foundation of China (No.623B2065).


\newpage
\bibliography{duplex}

\newpage

\appendix

\section{Transition Sentences}

To generate a sentence with coherent context, we utilize ChatGPT to rewrite the template below, which replaces \{sentence\_a\} and \{sentence\_b\} with one transition sentence and new content respectively.

\begin{tcolorbox}[boxsep=1mm, colback = white!25!white, colframe = black!75!black]
Fuse the two sentences smoothly and replace [topic] with the topic of sentence two.\\ 
\\
Sentence one\: "\{sentence\_a\}" \\
\\
Sentence two\: "\{sentence\_b\}" \\
\\
Give me your answer only, no other words. Give me your answer only, no other words.
\end{tcolorbox}

\subsection{Generation Termination Transition Sentences}
\label{sec:interruption_transition}

\begin{enumerate}
    \item <Empty string>
    \item I need to cut you off right now; this is urgent.
    \item Excuse me, I need to interject for a moment.
    \item Sorry to interrupt, but I have something important to add.
    \item Excuse me, may I interrupt for a moment?
    \item I'm sorry to break in, but there's something important I need to address.
    \item I apologize for interrupting, but I'd like to interject for a moment.
    \item I'm sorry to interrupt, but I have a quick point to make.
    \item I appreciate your input, but I need a moment of silence now.
    \item I'm sorry to interrupt, but I really need some quiet time to focus.
    \item Enough talking! I need you to be quiet now.
\end{enumerate}

\subsection{Regeneration Transition Sentences}
\label{sec:mistake_transition}

\begin{enumerate}
    \item I may not have expressed myself clearly. What I meant was [topic]
    \item I think there might be a bit of confusion. Let me clarify [topic]
    \item I appreciate your input, but I was hoping for more details on [topic]
    \item I think there might be a misunderstanding. What I'm really looking for is [topic]
    \item I may not have explained myself clearly. Let me rephrase the question. What are your thoughts on [topic]?
    \item Actually, the correct information is [topic]. Could you share your perspective on that?
    \item I'm a bit confused because what you mentioned contradicts the information I have. Can we go over this again?
    \item I'm sorry, but that information seems to be incorrect. Let me clarify the question, and please provide the accurate details regarding [topic].
    \item I'm sorry, but that's not accurate. The correct information is [topic]. It's essential to have the correct details for our discussion.
    \item I appreciate your effort in responding, but I think there might be a misunderstanding. What I intended to convey was [topic]. Let's revisit the topic to ensure we're on the same page.
    \item I see there might be some confusion. Let me clarify my point further to ensure we're on the same page. What I meant was [topic]. Can we discuss this to make sure we have a mutual understanding?
    \item There seems to be a misunderstanding. I meant [topic]. Let's align our understanding.
    \item No.
    \item Oh, No.
    \item No, you are wrong.
\end{enumerate}

\subsection{Dialogue Reset Transition Sentences}
\label{sec:topic_transition}

\begin{enumerate}
    \item <Empty string>
    \item That's interesting, and speaking of [topic], have you ever...?
    \item I was just thinking about [topic], what are your thoughts on that?
    \item That's fascinating! On a different note, have you ever thought about [topic]?
    \item I was just reading about [topic]. What are your thoughts on that?
    \item By the way, speaking of something else.
    \item That reminds me, have you heard about [topic]?
    \item Can we shift gears for a moment and talk about [topic]?
    \item I've been curious about [topic]. Have you ever considered it?
    \item I was thinking about [topic]. What are your thoughts on that?
    \item Now, shifting gears to a different subject, have you ever explored [topic]
    \item Moving on to a different topic, have you ever considered [topic]
    \item Changing the subject, have you ever thought about [topic]
    \item Switching gears, let's talk about [topic]
    \item On a different note, have you ever thought about [topic]
    \item Speaking of which, have you ever considered exploring [topic]
    \item Changing the subject, let's now delve into [topic]
    \item Shifting gears a bit, let's talk about [topic]
\end{enumerate}

\section{Questionnaire Details}

\subsection{Subject Instruction}

Before the experiment, we inform each participant of the subject instruction. The whole instruction is listed below:
\begin{enumerate}
    \item This experiment requires subjects to have conversations with chat models. The content does not involve any dangerous remarks or have an impact on the subjects' physical and mental health.
    \item This test includes two parts: chatting and interacting with the models and filling out the questionnaire.
    \item The models are voice input and output modes that support multiple rounds of dialogue. At the end of each dialogue, you can press the new conversation button to start a new round of conversation.
    \item The models are English models and only support English dialogue.
    \item There are two types of models, A and B. You must have at least 10 conversations with each model.
    \item We have included some questions to start the conversation, just for reference.
    \item This test mainly evaluates the performance of the two models in terms of response speed, human-likeness, faithfulness, factuality, and overall experience.
    \item After the chat, fill out the questionnaire.
\end{enumerate}

\subsection{Questionnaire}
\label{sec:appendix-survey-questions}

\begin{enumerate}
    \item Score the model's response speed to evaluate whether the model can respond to your request. 
    \item Score the faithfulness of the model's answers to evaluate whether the model understands your question, follows your instructions, and whether the answer is relevant to your chat topic.
    \item Score the factuality of the model's answers and evaluate whether the content of the answers is correct.
    \item Score the human-likeness of the model's answers and evaluate whether the conversation process between you and the model is close to the feeling of daily communication between people and whether the conversation process is smooth.
    \item Score the overall experience of the model.
\end{enumerate}

\section{Explanation of Ethical Concerns}
\label{app_human}
All participants are recruited from a partner company. Those experiments are conducted during their working hours and we do not pay them additionally.

In the human-evaluation experiment, we collect basic demographic characteristics information: gender, age, and educational qualification. We also collect their knowledge and usage of LLMs and voice assistants, which is tightly related to our research topic. As for the evaluation of the two chat models, we utilize their experience. The participants permit all those characteristics and experience information collection for research purposes only.

\section{Case Demonstration}
\begin{figure}[!htp]
    \centering
    \subfloat[Case A]{
    \includegraphics[width=0.7\linewidth]{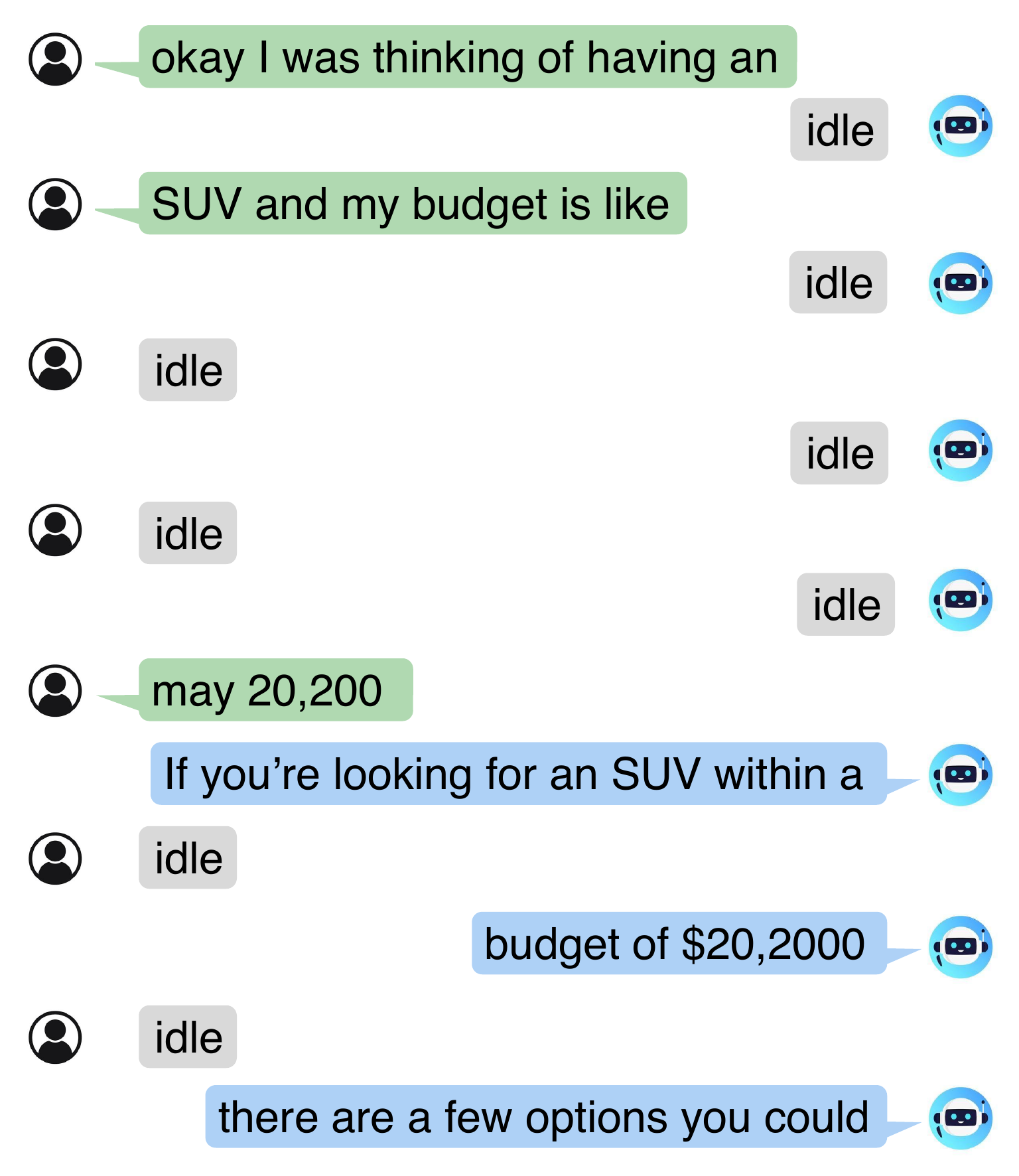}
    }\\
    \subfloat[Case B]{
    \includegraphics[width=0.7\linewidth]{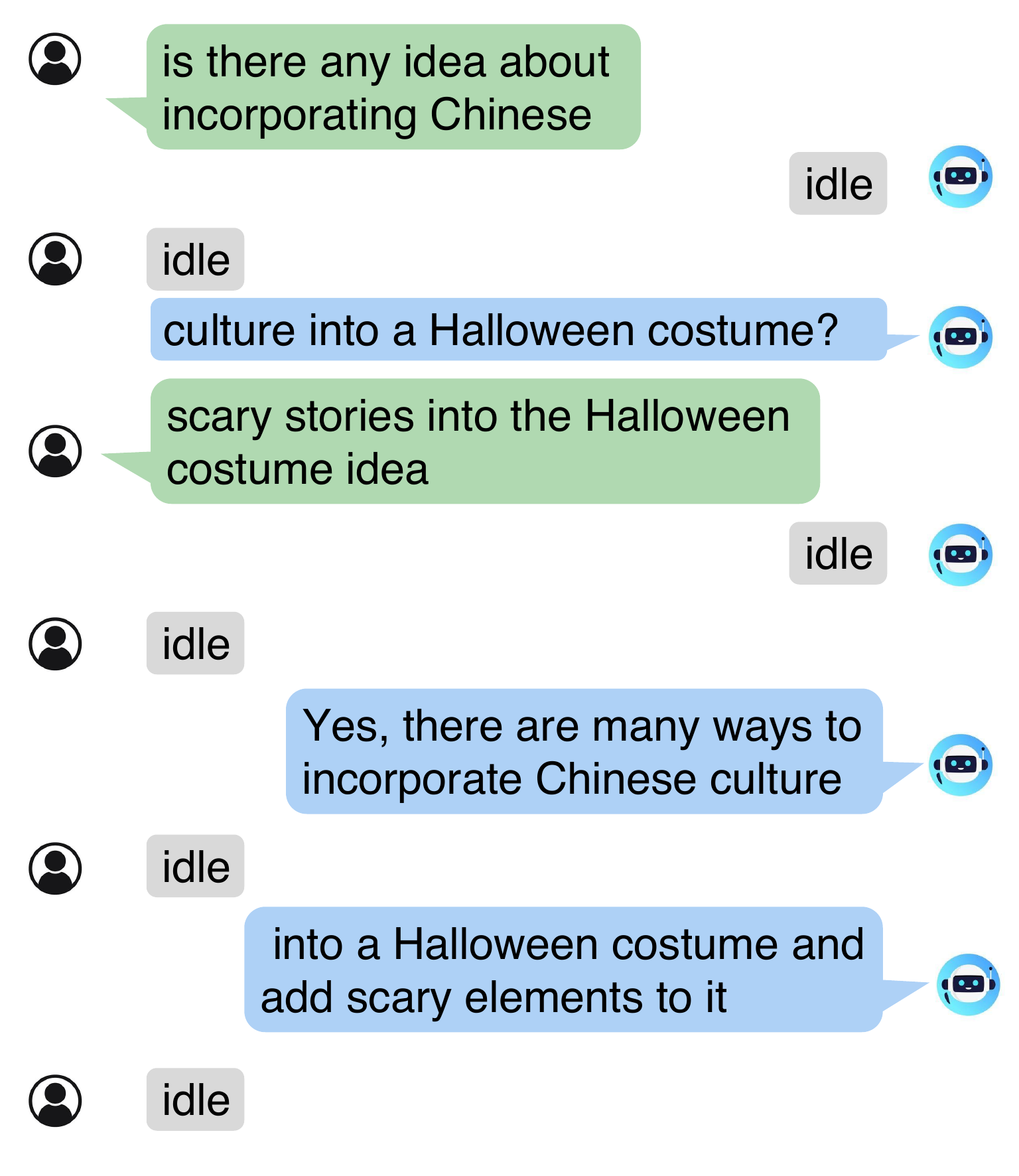}
    }\\
    \subfloat[Case C]{
    \includegraphics[width=0.7\linewidth]{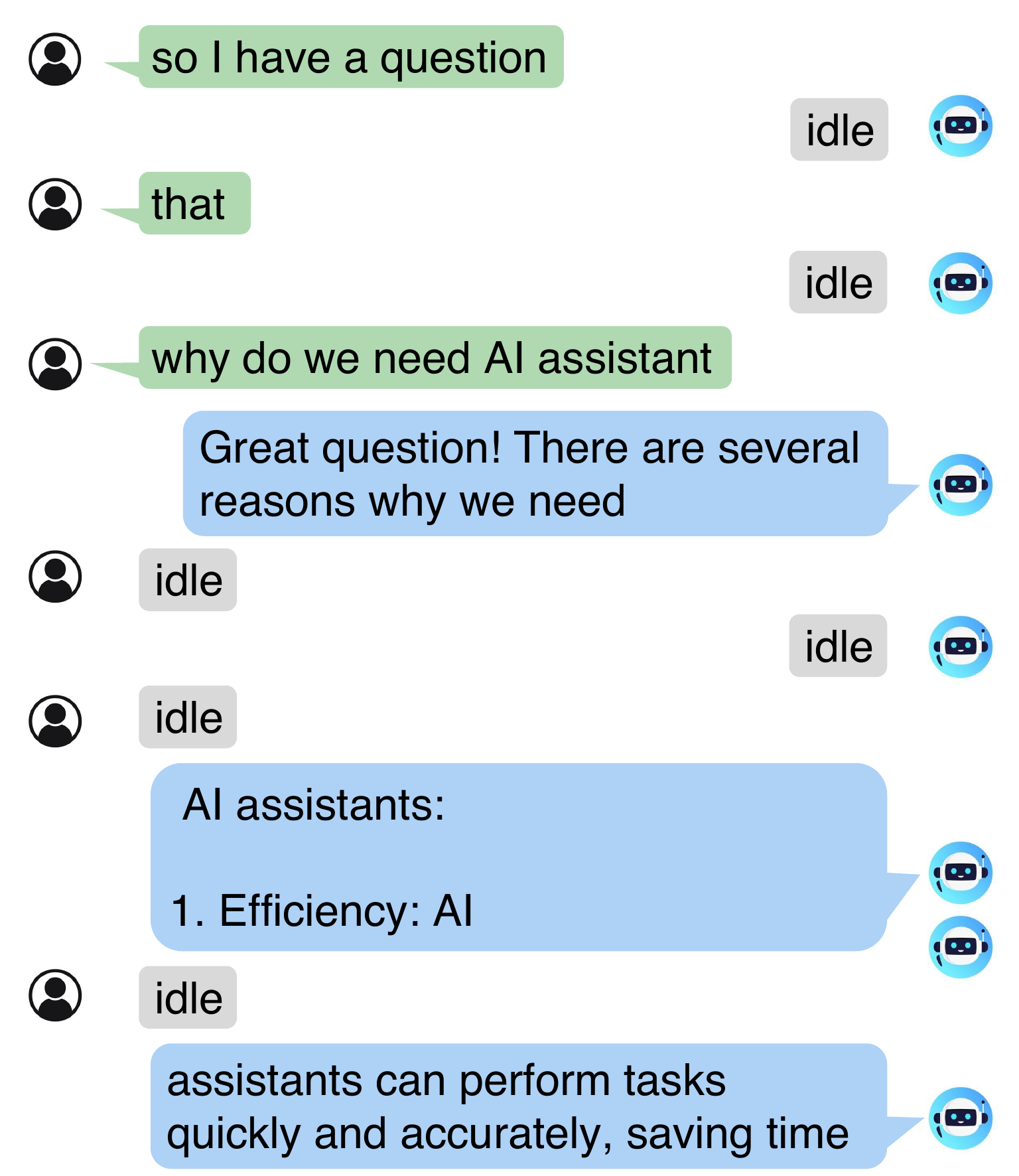}
    }
    \caption{User study cases.}
    \label{fig:case}
\end{figure}

Here are some cases of conversation segments between the MiniCPM-duplex and human users. In Figure~\ref{fig:case}, the duplex model generates a response until it obtains enough information from the user.

\end{document}